\documentclass[10pt,twocolumn,letterpaper]{article}

\usepackage{wacv}
\usepackage{times}
\usepackage{epsfig}
\usepackage{graphicx}
\usepackage{amsmath}
\usepackage{amssymb}

\usepackage{multirow}
\usepackage{multicol}

\usepackage{authblk}

\wacvfinalcopy 
\ifwacvfinal
\def\assignedStartPage{9876} 
\fi

\ifwacvfinal
\usepackage[breaklinks=true,bookmarks=false]{hyperref}
\else
\usepackage[pagebackref=true,breaklinks=true,colorlinks,bookmarks=false]{hyperref}
\fi

\ifwacvfinal
\else
\fi

\begin{document}

\title{Towards Precise Intra-camera Supervised Person Re-Identification}

\author[1]{Menglin Wang}
\author[2]{Baisheng Lai} 
\author[1]{Haokun Chen}
\author[2]{Jianqiang Huang}
\author[1]{Xiaojin Gong\thanks{The corresponding author.} \thanks{This work was supported by Major Scientific Research Project of Zhejiang Lab (No. 2019DB0ZX01).}}
\author[2]{Xian-Sheng Hua}
\affil[ ]{$^1$Zhejiang University, China, $^2$Damo Academy, Alibaba Group}
\affil[ ]{\tt\small {lynnwang6875@gmail.com; baisheng.lbs@alibaba-inc.com; chenhaokun@zju.edu.cn; jianqiang.hjq@alibaba-inc.com; gongxj@zju.edu.cn; huaxiansheng@gmail.com;}}


\maketitle

\begin{abstract}
Intra-camera supervision (ICS) for person re-identification (Re-ID) assumes that identity labels are independently annotated within each camera view and no inter-camera identity association is labeled. It is a new setting proposed recently to reduce the burden of annotation while expect to maintain desirable Re-ID performance. However, the lack of inter-camera labels makes the ICS Re-ID problem much more challenging than the fully supervised counterpart. By investigating the characteristics of ICS, this paper proposes jointly learned camera-specific non-parametric classifiers, together with a hybrid mining quintuplet loss, to perform intra-camera learning. Then, an inter-camera learning module consisting of a graph-based ID association step and a Re-ID model updating step is conducted. Extensive experiments on three large-scale Re-ID datasets show that our approach outperforms all existing ICS works by a great margin. Our approach performs even comparable to state-of-the-art fully supervised methods in two of the datasets.

\end{abstract}

\section{Introduction}
\label{sec:intro}
Person re-identification (Re-ID) is the task of matching images of the same person across disjoint cameras. Because of its significance in surveillance, this task has attracted broad research interest in recent years. Most previous works focus on fully supervised~\cite{kodirov2016person,sun2018beyond,zhang2019dsa,chen2019abd,luo2019trick} and unsupervised~\cite{deng2018similarity,chen2018deepa,zhong2019invariance,unsup_clustering,wu2019graph,yang2019patch} settings. The performance of supervised person Re-ID has been greatly improved due to the development of deep learning techniques. However, these methods need a large amount of full annotations that are expensive and time-consuming to obtain, leading to poor scalability in real-world deployments. Conversely, unsupervised methods require no annotations but their performance is still far from satisfactory.

\begin{figure}[t]
\centering
\includegraphics[width=0.45\textwidth]{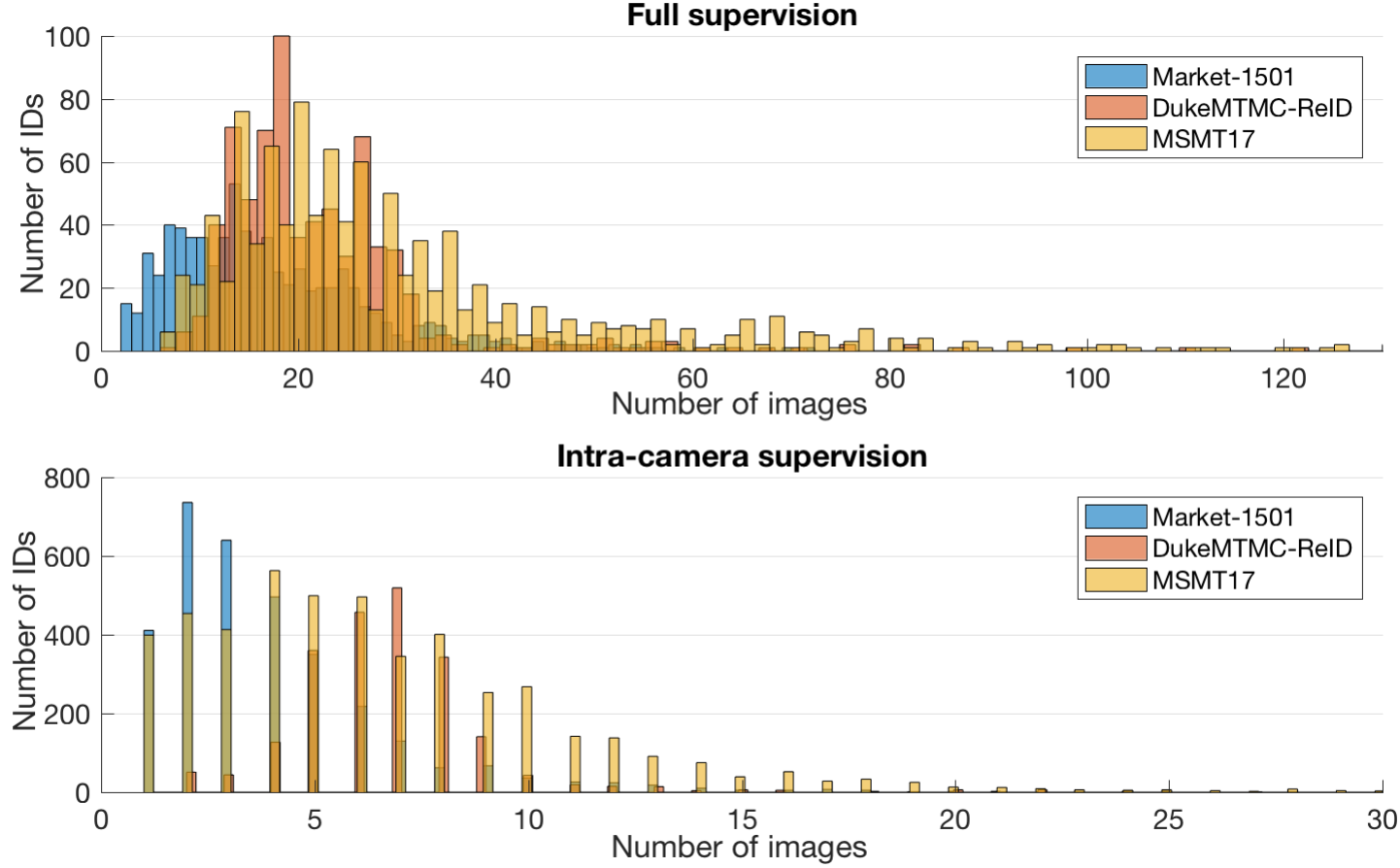}
\caption{Comparison of the ID histograms under full and intra-camera supervisions. The histograms are computed in three large-scale datasets: Market-1501~\cite{7410490}, DukeMTMC-reID~\cite{ristani2016performance,zheng2017unlabeled}, and MSMT17~\cite{wei2018person}.}
\label{fig_imageno}
\end{figure}

This paper aims to learn a person Re-ID model under intra-camera supervision (ICS), which is a new supervised setting proposed very recently~\cite{zhu2019intra,qi2019progressive}. It assumes that identity labels are independently annotated within each camera and no inter-camera identity association is labeled. Since the ID association across cameras is known as the most time-consuming step for manual annotation, ICS can greatly reduce annotation costs and make the Re-ID techniques more scalable. Nevertheless, the per-camera independent labeling nature brings up two challenges: 1) Although IDs within each camera are labeled, most IDs in ICS contain much less samples than those in full supervision, as shown in Figure~\ref{fig_imageno}. 2) The lack of inter-camera labels result in more difficulties when dealing with appearance variations in different cameras.


%


Therefore, how to properly exploit the labeled data within cameras and mine the unlabeled relations across cameras are two crucial problems in ICS. Existing works~\cite{zhu2019intra,zhu2020intra,qi2019progressive,qi2019intra} address these problems, respectively, from intra- and inter-camera learning perspectives. For intra-camera learning, MTML~\cite{zhu2019intra} and MATE~\cite{zhu2020intra} construct a multi-branch network that consists of a shared feature extraction backbone and multiple classification branches. Each branch learns a parametric softmax classifier for a specific camera view. However, the parametric classifiers could become ineffective when IDs contain only a couple of samples, which is a situation commonly occurred in ICS. Alternatively, PCSL~\cite{qi2019progressive} and ACAN~\cite{qi2019intra} perform the intra-camera learning by adopting the widely used triplet loss~\cite{hermans2017defense}, which mines the hardest positive and negative pairs within each training batch. The triplet loss takes only one negative sample each time and does not interact with other negative classes, leading to inferior performance and slow convergence~\cite{Sohn2016Npair}. These ineffective intra-camera learning consequently hurts the inter-camera learning part, leading to a Re-ID performance much lower than the counterparts in full supervision.

This paper aims to design a precise model for ICS person Re-ID. Considering that each camera has a variable number of IDs and most IDs have only few samples, we propose to jointly learn a set of camera-specific non-parametric classifiers for the intra-camera learning. These non-parametric classifiers, which are implemented with a shared feature extraction backbone and an external memory bank, are independent to the number of identity classes, enabling our network to entirely focus on the learning of discriminative features. Based on the memory bank, we also propose a quintuplet loss, which takes both in-batch samples and memory-stored ID centroids into account, to boost the intra-camera learning performance. These strategies help us to exploit the per-camera labeled data thoroughly and achieve a Re-ID model with considerable discrimination ability. Afterwards, an inter-camera learning step is followed that aims to improve the Re-ID model by mining ID relationships across cameras. To this end, we design a graph-based strategy for ID association and pseudo labeling. The pseudo labels are further utilized as ground-truth to train our network in a fully supervised manner and gain a better Re-ID model.



Although we address the ICS problem from intra- and inter-camera learning perspectives as existing ICS works~\cite{qi2019intra,qi2019progressive,zhu2019intra,zhu2020intra}, our approach is distinct in the learning strategies and achieves much higher Re-ID performance. Specifically,  our contributions are as follow:

1) We propose jointly learned camera-specific non-parametric classifiers and a quintuplet loss for intra-camera learning. These designs are customized not only for the characteristics of ICS but also for the memory assisted network architecture. They enable our intra-camera learning module to achieve a Re-ID performance better than existing ICS full models~\cite{qi2019intra,qi2019progressive,zhu2019intra,zhu2020intra} that consider both intra- and inter-camera learning parts.

2) Benefited from the high discriminative model obtained above, together with the proposed graph-based association strategy, we are able to get desirable pseudo labels in inter-camera learning. These labels enable us to effectively train our Re-ID model, which is built on the BNNeck augmented network architecture~\cite{luo2019trick}, in a fully supervised manner. Riding the wave of architectures successfully applied in the supervised Re-ID task also boosts the performance further.

3) Extensive experiments on three large-scale Re-ID datasets including Market-1501~\cite{7410490}, DukeMTMC-reID~\cite{ristani2016performance,zheng2017unlabeled}, and MSMT17~\cite{wei2018person}, show that the proposed approach outperforms previous ICS works by a great margin. Our performance is even comparable to fully supervised methods on the first two datasets.

\section{Related Work}
\subsection{Person Re-identification}
\textbf{Fully supervised person Re-ID} has made significant progress relying on the success of deep learning techniques. However, it remains to be an unsolved problem due to challenges arising from cluttered background, occlusion, as well as variations in illumination, pose, and viewpoint. Recent methods have exploited part-based features~\cite{sun2018beyond}, human semantics~\cite{zhang2019dsa}, attention mechanisms~\cite{chen2019abd}, or data generation~\cite{zheng2019dgnet} to tackle the challenges. These methods often lead to complex network architectures. An exceptional work is Bag of Tricks (BoT)~\cite{luo2019trick} that achieves the state-of-the-art performance by augmenting the baseline network with a simple BNNeck component. In this paper, we construct our network upon the BoT model, i.e. the BNNeck augmented ResNet-50~\cite{he2016deep}, to keep our backbone simple yet effective.

\textbf{Unsupervised person Re-ID} has attracted a lot of research interest in recent years. Existing methods can be roughly categorized into two groups. One is based on domain adaptation techniques~\cite{deng2018similarity,zhong2019invariance,ding2019adaptive} that transfer knowledge from labeled source domain to unlabeled target domain. The other group~\cite{unsup_clustering,lin2019aBottom,wu2019graph} is purely unsupervised that requires no external labeled data. Both types of the methods commonly perform a step to associate IDs across cameras via K-nearest neighbors~\cite{zhong2019invariance}, clustering~\cite{unsup_clustering,lin2019aBottom}, or graph~\cite{wu2019graph} based strategies. Our ICS work also adopts a graph-based ID association step. But, in contrast to use a graph-weighted loss~\cite{wu2019graph}, we formulate the association as a problem of finding connected components in a sparse graph.

\textbf{Semi-supervised person Re-ID} aims to learn a Re-ID model from both labeled and unlabeled data~\cite{yangqi2019semi}. Intra-camera supervision (ICS) is a special semi-supervised setting proposed recently. All existing ICS works address the problem by considering both intra- and inter-camera learning perspectives. For supervised intra-camera learning, PCSL~\cite{qi2019progressive} and ACAN~\cite{qi2019intra} take the extensively used triplet loss~\cite{hermans2017defense}, while MTML~\cite{zhu2019intra} and MATE~\cite{zhu2020intra} design a multi-branch structure to learn parametric classifiers. For inter-camera learning, ACAN~\cite{qi2019intra} develops a multi-camera adversarial learning approach to reduce the cross-camera data distribution discrepancy, PCSL~\cite{qi2019progressive} utilizes a soft-labeling scheme, MTML~\cite{zhu2019intra} and MATE~\cite{zhu2020intra} adopt multi-label learning strategies. Although we address the ICS problem from intra- and inter-camera learning perspectives as well, we propose distinct learning strategies in both parts and achieve much higher Re-ID performance.

\textbf{Difference from memory-assisted Re-ID methods.} Although memory assisted non-parametric classification has been adopted in unsupervised Re-ID~\cite{zhong2019invariance} and semi-supervised Re-ID tasks~\cite{yangqi2019semi}, our approach is designed specific to the intra-camera supervised setting. Specifically, in contrast to~\cite{zhong2019invariance,yangqi2019semi} that use a memory to store instance features and apply one classifier for all images, we construct multiple non-parametric classifiers, each of which corresponds to one camera and all of which are learned jointly. In addition, memory is also used in PCSL~\cite{qi2019progressive}, but PCSL uses ID centroids stored in the memory to weight a parametric classifier for cross-camera soft-labeling, while we use non-parametric classifiers for intra-camera learning.

\subsection{Parametric and Non-parametric Classifiers}
\textbf{Parametric classifiers} in this work refer to those implemented by fully connected (FC) layers in a deep neural network (DNN), usually trained with a cross-entropy softmax classification loss. Such classifiers have been successfully used in fully supervised person Re-ID~\cite{sun2018beyond,Zhai2019loss,luo2019trick}. However, they have the following drawbacks: 1) The training process pays much attention to learning parameters in the FC layers that are abandoned during inference for person Re-ID~\cite{Zhai2019loss}, making the learned feature representation less discriminative for test data. 2) The classifiers can not be learned effectively when there are a large number of identities while each identity only has a small number of instances~\cite{xiao2017memory}.

\textbf{Non-parametric classifiers} in a DNN are commonly implemented via a non-parametric variant of the softmax function. It is first proposed in a fully supervised person search task~\cite{xiao2017memory} and extensively adopted in unsupervised~\cite{wu2018memory,zhong2019invariance,zhong2019memory} and semi-supervised~\cite{chen2018memory,yangqi2019semi} learning. A common challenge in these tasks is that the number of classes is huge but each class contains only one or few samples. A DNN equipped with the non-parametric classifiers makes its parameters independent to the class number so that the training process entirely focus on the feature representation learning. This property benefits the Re-ID performance since a person is re-identified only based on the distance of features at the test time. Nevertheless, the non-parametric model may overfit more easily when training data is abundant enough~\cite{xiao2017memory}. Considering that each camera has a variable number of IDs and most IDs contain only a few of samples due to the per-camera labeling manner in ICS, we adopt non-parametric classifiers for our intra-camera learning. While in inter-camera learning where IDs contain more samples after association, we choose the parametric variant.

\begin{figure*}[t]
\centering
\includegraphics[width=0.96\textwidth]{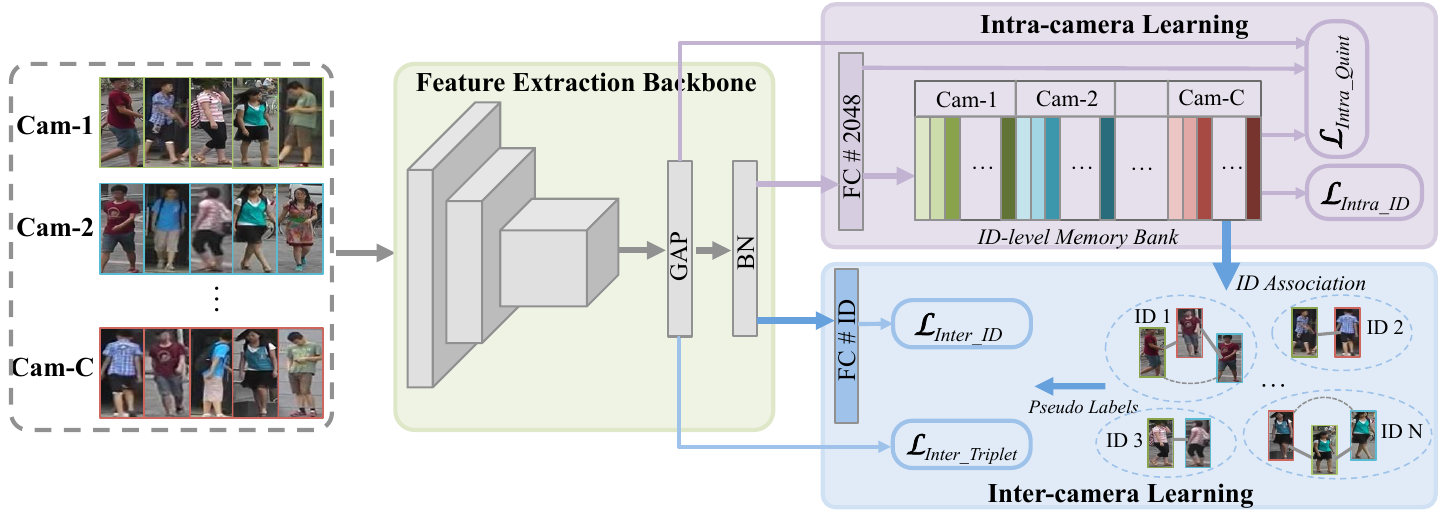}
\caption{An overview framework of the proposed method. It consists of a feature extraction backbone, together with an intra-camera learning branch and an inter-camera learning branch. In intra-camera learning, features are classified via camera-specific non-parametric classifiers implemented via an external memory bank, optimized by $\mathcal{L}_{Intra\_ID}$ and $\mathcal{L}_{Intra\_Quint}$. In inter-camera learning, IDs are associated across cameras and pseudo labeled, which are further used to update the Re-ID model by optimizing $\mathcal{L}_{Inter\_ID}$ and $\mathcal{L}_{Inter\_Triplet}$.}
\label{framework_pic}
\end{figure*}

\section{The Proposed Method}
The intra-camera supervision assumes that identity labels are independently annotated within each camera view and no inter-camera identity association is provided. Suppose there are $C$ cameras in a dataset. We denote the set of the $c$-th camera by $\mathcal{D}_c = \{(x_i, y_i, c_i)\}_{i=1}^{M_c}$, in which image $x_i$ is annotated with an identity label $y_i \in \{1, \cdots, N_c\}$ and a camera label $c_i = c \in \{1, \cdots, C\}$. $M_c$ and $N_c$ are, respectively, the number of total images and IDs in this camera view. $N = \sum_{c=1}^{C} N_c$ is  the total ID number directly accumulated over all cameras. It should be noted that the identities in different cameras are partially overlapped. That is, a same person may appear in two or more camera views, but it could be assigned with different IDs due to the per-camera independent labeling manner. Given this training set $\mathcal{D}=\bigcup_{c=1}^C \mathcal{D}_c$, we aim to learn a Re-ID model that can well discriminate both intra- and inter-camera identities.

To this end, we develop our method from both intra- and inter-camera learning perspectives. The overall framework is shown in Figure~\ref{framework_pic}. An image is first fed into a backbone network for feature extraction. The extracted feature goes through an additional feature embedding layer and then classified by camera-specific non-parametric classifiers that are implemented via an external memory bank, together with an ID classification loss and a quintuplet loss, for intra-camera learning. The memory bank stores the centroid feature of each ID, which is of moderate discrimination ability after intra-camera learning. Then, the ID centroids are used for ID association and pseudo labeling across cameras. In inter-camera learning, the same backbone is adopted to extract features, along with a classifier parameterized by a FC layer to classify images into their pseudo identity classes.

\subsection{Intra-camera Learning}
When considering the Re-ID problem within an individual camera, it can be treated as a fully supervised classification task. Therefore, it is reasonable to formulate the intra-camera learning as a fully supervised multi-task classification problem that learns multiple per-camera ID classifiers jointly. However, the number of IDs in each camera varies a lot and most IDs have only several samples, making parametric classifiers implemented via multi-branch fully-connected architectures~\cite{zhu2019intra,zhu2020intra} ineffective. Thus, we opt to use non-parametric variants in this work. 

We design a number of non-parametric classifiers, each of which performs the fully supervised ID classification within one camera view. All non-parametric classifiers share a feature extraction backbone and an external memory bank so that the classifiers are learned jointly. Each classifier, termed as a \textit{camera-specific non-parametric classifier}, optimizes an ID classification loss that pulls an image close to the centroid of its labeled ID while pushes away from the centroids of all other IDs in the same camera. By considering the distances between samples and centroids, the ID loss is good at separating a majority of samples, but may still fail for hard ones. Regarding to this reason, we also propose a hybrid mining quintuplet loss to improve the inter-class separability. The details are introduced in the followings.

\subsubsection{Camera-specific Non-parametric Classifiers}
\label{sec_intracls}
As illustrated in Figure~\ref{framework_pic}, our network consists of a feature extraction backbone, a FC embedding layer, together with an external memory bank. Based upon this network structure, we design camera-specific non-parametric classifiers to perform the classification tasks within each camera view.

Formally, when an image is input, the FC embedding layer outputs a $d$-dimensional feature vector ($d=2048$ in our work). The memory bank $\mathcal{K}\in R^{d\times N}$ stores the up-to-date features of all accumulated IDs and each column corresponds to one ID. During back-propagation, the memory bank is updated by
\begin{equation}
	\mathcal{K}[j] \leftarrow \mu \mathcal{K}[j] + (1 - \mu) f(x_i),
	\label{eq:mu}
\end{equation}
where $\mathcal{K}[j]$ is the $j$-th column of the memory. $f(x_i)$ is a $L_2$ normalized feature extracted from image $x_i$ that belongs to the $j$-th ID. $\mu \in [0,1]$ is an updating rate. After each update, $\mathcal{K}[j]$ is scaled to having unit $L_2$ norm. The updated feature in each column can be interpreted as the centroid of an identity class in the feature space, which is a $d$-dimensional unit hypersphere.

Given the image $x_i$, together with its annotated intra-camera identity label $y_i$ and camera label $c_i$. The corresponding global ID index $j$ is obtained by $j = A + y_i$, where $A = \sum_{k=1}^{c_i-1} N_k$ is the total ID number accumulated from the first to the $c_i-1$-th camera view. Then, the probability of classifying $x_i$ into the $j$-th ID is defined by a non-parametric softmax function
\begin{equation}
p(j|x_i) = \frac{exp(\mathcal{K}[j]^T f(x_i)/\tau)}{\sum_{k=A+1}^{A+N_{c_i}} exp(\mathcal{K}[k]^T f(x_i)/\tau)},
\label{eq_memory_prob}
\end{equation}
where $\tau$ is the temperature controlling the smoothness of probability distribution.

Note that the non-parametric classifier defined above is camera-specific, because the sum in the denominator is over the IDs within the same camera view only. In contrast to existing works~\cite{xiao2017memory,wu2018memory,zhong2019invariance,ding2019adaptive} that considers all entries in their memory bank, our formulation only takes those belonging to the same camera into account while ignores the IDs in all other cameras. Thus, each non-parametric classifier is responsible for the classification task in a specific camera. 

To jointly learn all the camera-specific identity classification tasks together, the objective, termed as the \textit{intra-camera ID loss}, is formulated to minimize the negative log-likelihood of training images in all cameras. That is, 
\begin{equation}
\mathcal{L}_{Intra\_ID}=-\sum_{c=1}^{C}\left(\frac{1}{|\mathcal{D}_c|}\sum_{(x_i, y_i, c_i) \in \mathcal{D}_c} \log p(j|x_i)\right),
\label{eq_memory_loss}
\end{equation}
where $\frac{1}{|\mathcal{D}_c|}$ is a normalization coefficient placed to balance the variable number of images in different cameras.

\subsubsection{A Hybrid Mining Quintuplet Loss}
\begin{figure}[t]
\centering
\includegraphics[width=0.48\textwidth]{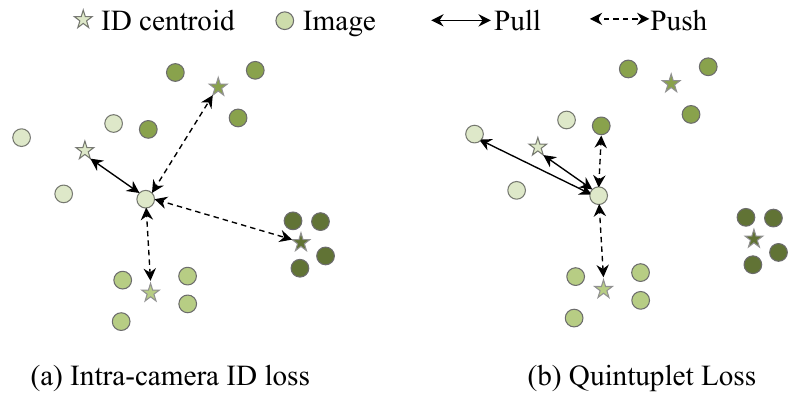}
\caption{Illustration of intra-camera ID loss and quintuplet loss.}
\label{fig_loss}
\end{figure}

The intra-camera ID loss introduced above in essence pulls a sample close to its centroid and meanwhile pushes it away from the centroids of all other IDs within the same camera, as shown in Figure~\ref{fig_loss}(a). This loss classifies most samples well, but may be failed for hard examples. To improve the inter-class separability and intra-class compactness further, we propose a quintuplet loss as a supplement. Our loss is constructed based upon the widely used triplet loss~\cite{hermans2017defense}. In addition to the triplet loss that mines hard positive and negative samples locally within each mini-batch, the quintuplet loss also incorporates the hard pairs stored in the global memory bank. It pulls an anchor close to both its centroid and the associated hard positive sample, and pushes away from both the hard negative sample and ID centroid, as illustrated in Figure~\ref{fig_loss}(b).


Mathematically, in each mini-batch, we randomly select $P$ identities and $K$ instances of each identity, as the common practice~\cite{hermans2017defense}. For each anchor image $x_a$, we design a hybrid mining scheme that selects two instances and two identity centroids to form a quintuplet. The positive and negative instances are sampled to be the hardest ones within a mini-batch. In addition, we choose the positive ID centroid and the nearest negative ID centroid from the memory bank. Note that all instances and centroids are selected from the same camera as the anchor. Then, the intra-camera quintuplet loss is defined as follows:
\begin{equation}
\small{
\begin{aligned}
& \mathcal{L}_{Intra\_Quint}= \sum_{i=1}^{P}\sum_{a=1}^{K} \Big[ m_1 + \mathop{max}\limits_{p=1,...,K}{||g(x_a^{i})-g(x_p^{i})||} \\
&\ \ \ \ \ \ \ \ \ \ \ \  -\mathop{min}\limits_{\substack{n=1,...,K; j=1,...,P; j\neq i; c_a^i=c_n^j}}{||g(x_a^{i})-g(x_n^{j})||} \Big]_+\\
&\ \ \ \ \ \ \ \ \ \ \ \  + \Big[ m_2 +||f(x_a^i) - \mathcal{K}[A+y_a^i])||\\
&\ \ \ \ \ \ \ \ \ \ \ \  -\mathop{min}\limits_{\substack{j=1,...,N_{c_a^i}; j\neq y_a^i}}{||f(x_a^i)-\mathcal{K}[A+j]||} \Big]_+,
\end{aligned}
}
\label{eq_quintuplet_loss}
\end{equation}
where $m_1$ and $m_2$ are two margins. $g(x_a)$, $g(x_p)$, and $g(x_n)$ are, respectively, the features of the anchor, positive and negative instances output from the global average pooling (GAP) layer in the backbone. $f(x_a)$ is the anchor's feature produced from the FC embedding layer as above-mentioned. Taking features from different layers is inspired by the BNNeck structure in~\cite{luo2019trick} and shown to be effective in our experiments. In addition, $A+y_a$ is the global ID index in $\mathcal{K}$ given the intra-camera ID $y_a$, $[\cdot]_+ = max(0, \cdot)$, and $||\cdot||$ is the Euclidean distance.

\subsubsection{The Loss for Intra-camera Learning}
In summary, the loss for intra-camera learning is the sum of the intra-camera ID loss and the quintuplet loss:
\begin{equation}
\mathcal{L}_{Intra}=\mathcal{L}_{Intra\_ID} + \mathcal{L}_{Intra\_Quint}.
\label{eq_intra_total_loss}
\end{equation}

\subsection{Inter-camera Learning}
With the proposed camera-specific non-parametric ID classification loss and quintuplet loss, our Re-ID model achieves outstanding discrimination ability within cameras. Meanwhile, the model also gains certain capability in discriminating IDs across cameras through the joint learning of all per-camera classification tasks. However, due to the lack of explicit inter-camera correlations, the model is still weak at coping with variations in different cameras. To address this problem, we perform an inter-camera learning that consists of a cross-camera ID association step and a Re-ID model updating step.


\subsubsection{Cross-camera ID Association}
We formulate the cross-camera ID association task as a problem of finding connected components in a graph. We construct the graph based on two observations: 1) The more similar two IDs are, the more likely they are to be the same person. 2) Under the intra-camera supervised setting, each ID has no positive matches within the same camera, and at most one positive match in any associated camera. According to these observations, we construct an undirected graph $\mathcal{G}=\left\langle \mathcal{V}, \mathcal{E}\right\rangle$ for association, where the vertex set $\mathcal{V}$ denotes the IDs accumulated over all cameras and the edge set $E=\{e(i,j)\}$ indicates whether the $i$-th ID and the $j$-th ID is a positive pair or not. The edge $e(i,j)$ is defined by
\begin{equation}
e(i,j) = \left\{\begin{array}{l}
						1, \quad dist(i,j)<T \ \wedge \ c(i)\neq c(j)  \\
						\ \ \ \ \ \ \   \wedge \ i\in \mathcal{N}_1(j, c(i)) \ \wedge \ j\in \mathcal{N}_1(i, c(j)); \\
						0, \quad otherwise. \end{array} \right.
\end{equation}
Here, $dist(i,j)=||\mathcal{K}[i] - \mathcal{K}[j]||$ is the Euclidean distance between two ID centroids stored in the memory bank, and $i$, $j \in \{1,\cdots ,N\}$. $T$ is a threshold taken by ascendingly sorting the distances of ID pairs and choosing the $S$-th distance value, implying to choose Top $S$ similar pairs. $c(i)$ represents the camera that the $i$-th ID belongs to. $\mathcal{N}_1(j, c(i))$ is the 1-nearest neighbor of ID $j$ in camera $c(i)$ and likewise for $\mathcal{N}_1(i, c(j))$. The last two conditions require $i$ and $j$ to be the reciprocal nearest neighbor of each other. 


The constructed graph is sparsely connected. Therefore, we adopt a union-find algorithm \cite{patwary2010experiments} to find all connected components in the graph. All IDs within each component are associated and assigned with a same pseudo label. The pseudo labels are further used to update the Re-ID model.  


\subsubsection{Re-ID Model Updating}
\label{sec:model_updating}
Taking all images and their pseudo labels, we treat the Re-ID task as a fully supervised problem and adopt the simple yet effective architecture in BoT~\cite{luo2019trick} to update the Re-ID model. Specifically, based upon the feature extraction backbone learned at our intra-camera learning stage, we append a FC layer to learn a parametric ID classifier, as shown in Figure~\ref{framework_pic}. The model is trained with the extensively used cross-entropy loss with the label smoothing scheme, termed as an inter-camera ID loss $\mathcal{L}_{Inter\_ID}$ in our work, together with a batch-hard triplet loss $\mathcal{L}_{Inter\_Triplet}$ applied to the features output from the GAP layer. Therefore, the total loss for inter-camera learning is:
\begin{equation}
\mathcal{L}_{Inter} = \mathcal{L}_{Inter\_ID} + \mathcal{L}_{Inter\_Triplet}.
\end{equation}

Note that the above-mentioned inter-camera ID loss is using a parametric Softmax. Analogous to the intra-camera learning, we can also implement it via the memory-assisted non-parametric Softmax. However, the non-parametric version is prone to overfit when training data become abundant after ID association, leading to an inferior performance. 


\section{Experiments}
\subsection{Experiment Setting}
\textbf{Datasets and Evaluation Metrics. }
We evaluate the proposed method on three large-scale datasets: Market-1501 \cite{7410490}, DukeMTMC-reID \cite{ristani2016performance,zheng2017unlabeled}, and MSMT17 \cite{wei2018person}. The numbers of cameras, IDs, and images contained in each dataset are reported in Table~\ref{dataset_statistic_table}. To simulate the ICS setting, we generate intra-camera identity labels based on the provided full annotations. Table~\ref{dataset_statistic_table} also lists the accumulated total identity number under intra-camera supervision ($\#ID_{ICS}$), the averaged image-per-person (IP) value, and the averaged image-per-camera-per-person (ICP) value. For performance evaluation, we adopt the Cumulative Matching Characteristic (CMC) and mean Average Precision (mAP), as the common practice.
\begin{table}[ht]
\centering
\scalebox{0.75}{
\begin{tabular}{c|c|c|c|c|c|c}
\hline  
Dataset & $\#Cam$  & $\#ID$  &  $\#Img$  & $\#ID_{ICS}$ & IP   & ICP \\
\hline
Market-1501             & 6 & 751 & 12,936 & 3,262 & 17.23  &  3.97 \\
DukeMTMC-reID      & 8 & 702 & 16,522 & 2,196 &  23.54  &  7.52 \\
MSMT17                   & 15 & 1,041 & 32,621 & 4,821 & 31.34  &  6.77 \\
\hline
\end{tabular}
}
\caption{\scriptsize {Dataset statistics. $\#Cam$, $\#ID$, $\#Img$, and $\#ID_{ICS}$ are the number of cameras, IDs, images, and accumulated IDs under ICS, respectively. IP is the averaged image-per-person value, ICP is the averaged image-per-camera-per-person value.}}
\label{dataset_statistic_table}
\end{table}

\textbf{Implementation details. }
We adopt the ImageNet~\cite{krizhevsky2012imagenet}-pretrained ResNet-50~\cite{he2016deep} as our feature extraction backbone. Following~\cite{luo2019trick}, we remove the last spatial down-sampling operation in the backbone to increase the size of feature maps, and add a batch normalization (BN) layer after GAP. During both intra- and inter-camera learning, images are resized to $256\times 128$. Random flipping, cropping, and erasing are performed as data augmentation. The initial learning rate is $0.00035$ and divided by $10$ after $40$ and $70$ epochs. We choose Adam~\cite{Kingma2014} as the optimizer with weight decay $0.0005$. Training batch size is $64$, with randomly selected $16$ IDs and $4$ images for each ID. In intra-camera learning, images within each mini-batch are sampled according to the per-camera labels, and the total number of epochs is $50$. Temperature $\tau$ in Equation (\ref{eq_memory_prob}) is set to $0.067$ (i.e.$1/15$). The margins $m_1$ and $m_2$ in Equation (\ref{eq_quintuplet_loss}) are both empirically set to $0.3$. In inter-camera learning, images in each mini-batch are sampled according to the generated pseudo ID labels. The total number of epochs is $120$. In graph construction, we select Top $S$ similar pairs as candidates for edge linking, where $S$ is empirically set to be the same number as $\#ID_{ICS}$ listed in Table~\ref{dataset_statistic_table}. Each experiment runs 5 times and the averaged performance is reported.

\subsection{Ablation Study}

\subsubsection{Effectiveness of The Intra-camera Learning Part}

\textbf{Effectiveness of The Camera-specific Non-parametric Classifiers. } 
First, we are curious about how well the camera-specific non-parametric classifiers perform. Therefore, three model variants are investigated, including $\mathcal{M}_1$: a multi-branch network structure~\cite{zhu2019intra}, in which each branch uses a classifier parameterized by a FC layer and optimized with a cross-entropy ID loss, for intra-camera learning; $\mathcal{M}_2$: a non-parametric classifier but not camera-specific, that is, any image can be classified into all accumulated ID classes; $\mathcal{M}_3$: the proposed camera-specific non-parametric classifiers with the intra-camera ID loss only.

\begin{figure*}[t]
\begin{minipage}[t]{0.32\textwidth}
\centering
\includegraphics[width=1\textwidth]{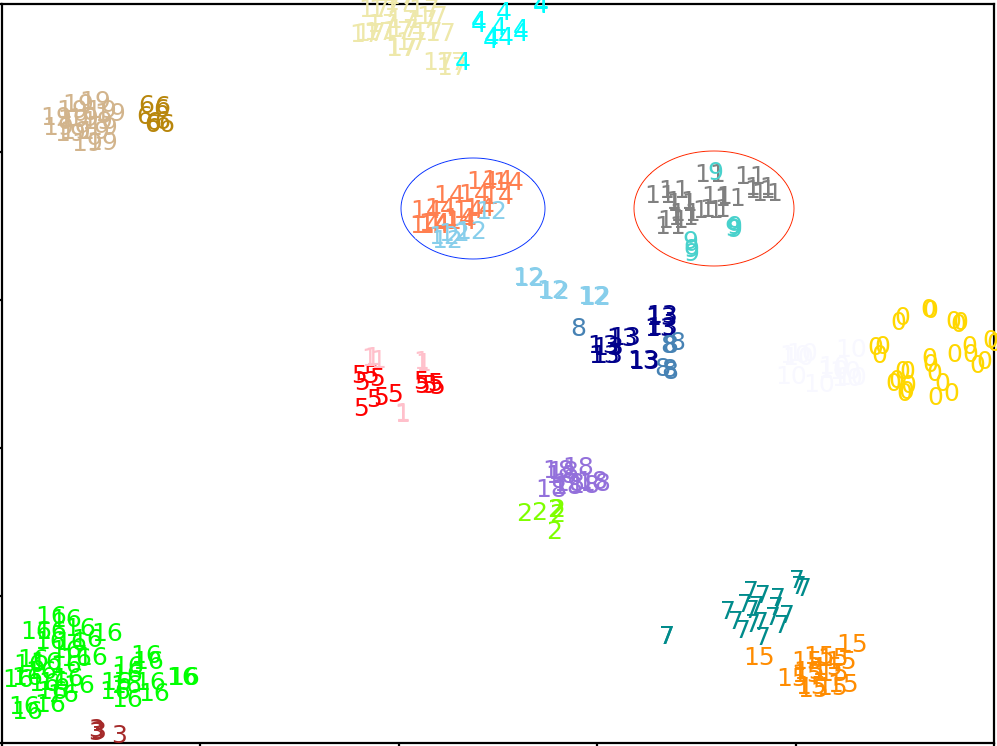}
\end{minipage}\hfill
\begin{minipage}[t]{0.32\textwidth}
\centering
\includegraphics[width=1\textwidth]{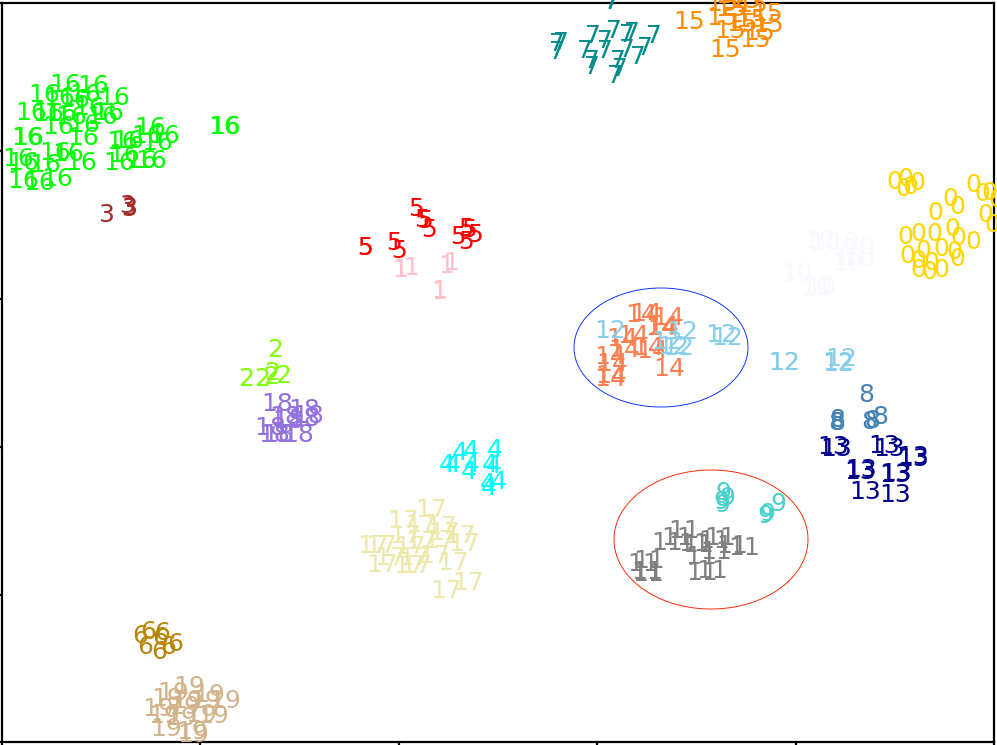}
\end{minipage}\hfill
\begin{minipage}[t]{0.32\textwidth}
\centering
\includegraphics[width=1\textwidth]{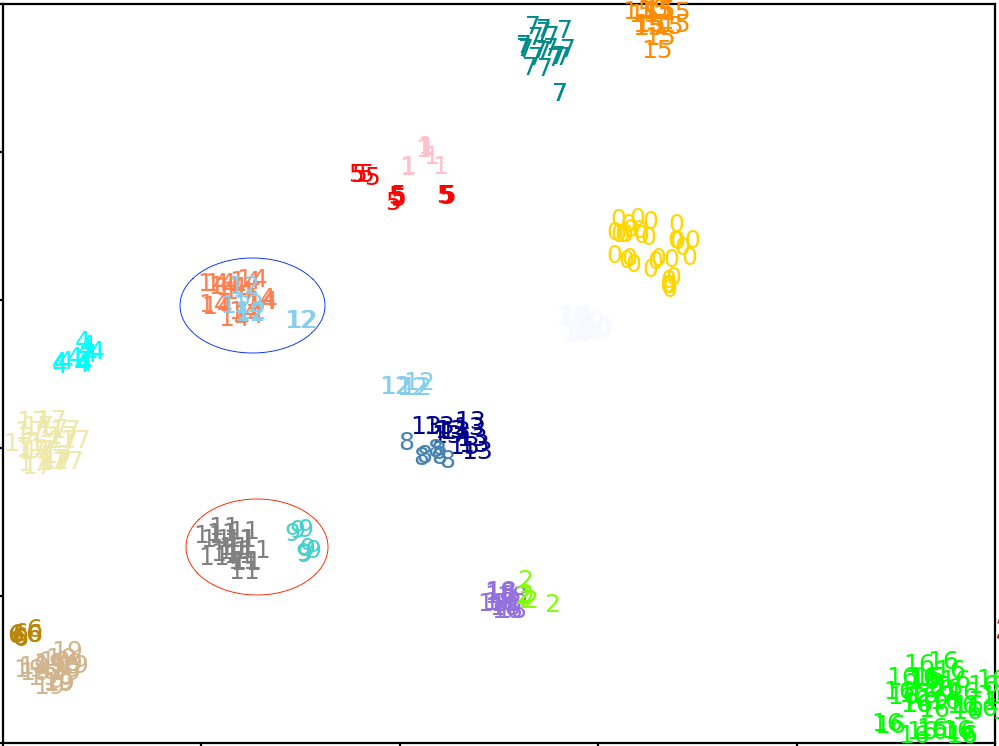}
\end{minipage}
\\
\begin{minipage}[t]{0.24\textwidth}
\centering
\includegraphics[width=1\textwidth]{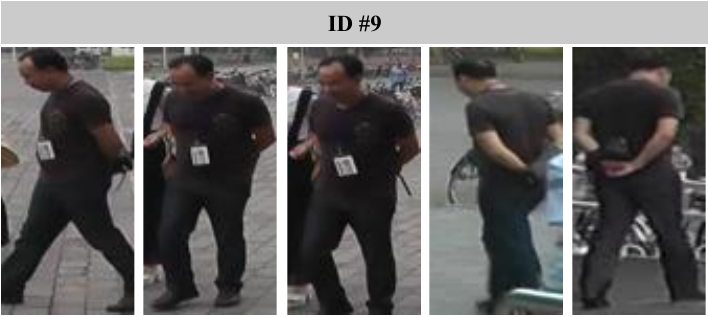}
\end{minipage}\hfill
\begin{minipage}[t]{0.24\textwidth}
\centering
\includegraphics[width=1\textwidth]{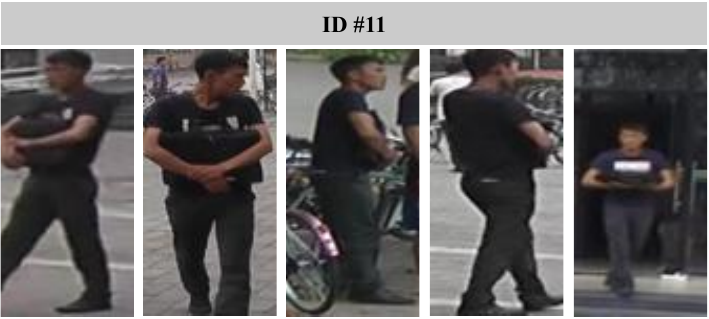}
\end{minipage}\hfill
\begin{minipage}[t]{0.24\textwidth}
\centering
\includegraphics[width=1\textwidth]{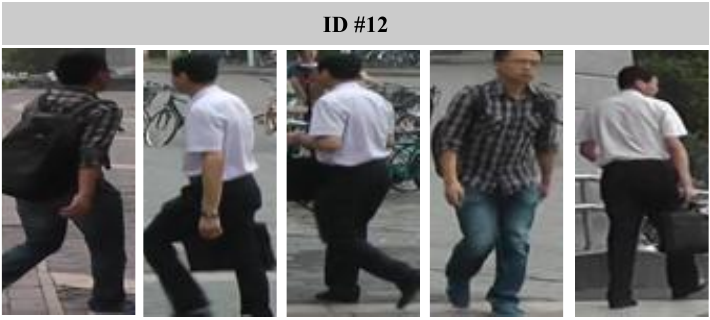}
\end{minipage}\hfill
\begin{minipage}[t]{0.24\textwidth}
\centering
\includegraphics[width=1\textwidth]{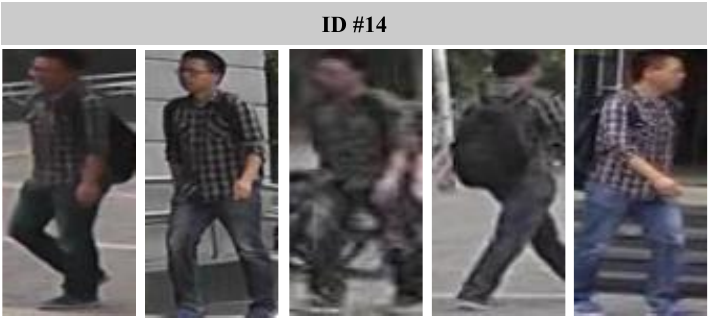}
\end{minipage}
\caption{T-SNE visualization of the features learned by $\mathcal{M}_3$, $\mathcal{M}_5$, and $\mathcal{M}_7$, which are respectively shown from left to right in the first row. Typical example images from IDs $\#9, \#11, \#12, \#14$ are presented in the bottom row.}
\label{tsne_visualization}
\end{figure*}

Table~\ref{ablation_table} presents the comparison results in terms of mAP(\%) and Rank-1(\%). From the results we observe that the camera-specific non-parametric classifiers ($\mathcal{M}_3$) outperform the camera-agnostic counterpart ($\mathcal{M}_2$) by a great margin, showing that the camera-specific scheme plays an important role. In addition, the camera-specific non-parametric classifiers ($\mathcal{M}_3$) perform consistently better than the parametric counterpart ($\mathcal{M}_1$) on all datasets. Especially, it improves the performance by a significant margin (mAP +14.2\% and Rank-1 +9.3\%) on Market1501 which has a smaller ICP value than the other two datasets, as reported in Table~\ref{dataset_statistic_table} and Fig.~\ref{fig_imageno}. It indicates that the non-parametric classifiers are superior to the parametric counterpart especially when identity classes contain fewer examples. 

\newcommand{\tabincell}[2]{\begin{tabular}{@{}#1@{}}#2\end{tabular}}
\begin{table}[htp]
\centering
\scalebox{0.85}{
\begin{tabular}{c|cc|cc|cc}
\hline  
\multirow{2}{*}{Models} &  \multicolumn{2}{c|}{Market1501} & \multicolumn{2}{c|}{DukeMTMC-ReID}& \multicolumn{2}{c}{MSMT17} \\
\cline{2-7} & Rank-1 & mAP  & Rank-1  & mAP    & Rank-1  & mAP \\ 
\hline
$\mathcal{M}_1$                  & 76.8  & 55.0           & 75.3  & 58.9       & 50.7 & 25.1 \\
$\mathcal{M}_2$                        &45.5 & 30.7              & 33.3 & 29.1        & 10.4 & 6.0 \\
$\mathbf{\mathcal{M}_3}$                        & \textbf{86.1}  &  \textbf{69.2}          & \textbf{78.0} & \textbf{61.9}        & \textbf{52.1}& \textbf{25.7}\\ \hline
$\mathcal{M}_4$                        & 86.8  &  71.9          &79.1  & 64.0        & 54.3 & 28.1 \\
$\mathbf{\mathcal{M}_5}$                        & \textbf{87.5}  &  \textbf{72.3}         & \textbf{79.7}  & \textbf{64.7}         & \textbf{55.5} & \textbf{28.9}\\ \hline
$\mathbf{\mathcal{M}_6}$  						 & 91.5 & 79.7             & 80.6 & 65.5         & 49.7 & 23.4  \\
$\mathbf{\mathcal{M}_7}$                       & \textbf{93.1} & \textbf{83.6}             & \textbf{83.6} & \textbf{72.0}          & \textbf{57.7} & \textbf{31.3} \\ \hline
$\mathcal{M}_8$                        & 94.1  & 85.9           & 87.4  &77.2         & 75.3  & 52.2\\
\hline
\end{tabular}
}
\caption{Comparison of the different model variants. $\mathcal{M}_1$-$\mathcal{M}_5$ are intra-camera model variants, in which $\mathcal{M}_1$ adopts multi-branch parametric classifiers, $\mathcal{M}_2$ adopts camera-agnostic non-parametric classifiers, and $\mathcal{M}_3$ is our proposed camera-specific non-parametric classifiers. $\mathcal{M}_4$ is $\mathcal{M}_3$ with an additional triplet loss and $\mathcal{M}_5$ is $\mathcal{M}_3$ with the proposed quintuplet loss. $\mathcal{M}_6$ and $\mathcal{M}_7$ are, respectively, adopting a non-parametric and a parametric inter-camera learning part based upon $\mathcal{M}_5$. $\mathcal{M}_8$ is a fully supervised version.}
\label{ablation_table}
\end{table}

\textbf{Effectiveness of The Quintuplet Loss. }
When validating the effectiveness of the proposed quintuplet loss, we investigate two model variants, which are $\mathcal{M}_4$: the camera-specific non-parametric classifiers with the intra-camera ID loss and a batch-hard triplet loss~\cite{hermans2017defense}; and $\mathcal{M}_5$: the camera-specific non-parametric classifiers with the intra-camera ID loss and the proposed quintuplet loss. From the results reported in Table~\ref{ablation_table}, we observe that both $\mathcal{M}_4$ and $\mathcal{M}_5$ gain considerable improvements when compared to $\mathcal{M}_3$ that does not consider hard samples. Moreover, the model using the quintuplet loss ($\mathcal{M}_5$) performs consistently better than the one using the triplet loss ($\mathcal{M}_4$), demonstrating that it is effective to leverage information from both the local mini-batch and the global memory bank.

\begin{table}[h]
\centering
\scalebox{0.9}{
\begin{tabular}{cc|cc}
\hline  
\multicolumn{2}{c|}{Market1501} & \multicolumn{2}{c}{DukeMTMC-ReID} \\
\hline
Precision  & Recall  & Precision & Recall   \\ 
\hline
96.4$\%$   & 75.9$\%$ & 90.1$\%$  & 74.3$\%$ \\
\hline
\end{tabular}
}
\caption{Precision and recall of ID pairs associated by our inter-camera association strategy.}
\label{table_prec_rec}
\end{table}

\begin{table*}[h]
\centering
\scalebox{0.95}{
\begin{tabular}{c|c|cccc|cccc|cccc}
\hline  
\multirow{2}{*}{Methods} & \multirow{2}{*}{Reference} &  \multicolumn{4}{c| }{Market1501} & \multicolumn{4}{c|}{DukeMTMC-ReID}  & \multicolumn{4}{c}{MSMT17}  \\
\cline{3-14}&  & R1 & R5 & R10 & mAP   & R1 & R5 & R10  & mAP   & R1 & R5 & R10 & mAP \\ 
\hline
\multicolumn{13}{l}{\textbf{Fully supervised}} \\
OSNet~\cite{zhou2019osnet}   & ICCV19  & 94.8 &- &- & 84.9                  &88.6 &- &- &73.5            &78.7 &-  &- &52.9\\ 
DGNet~\cite{zheng2019dgnet}  & CVPR19  &94.8 &- &- & 86.0                 &86.6 &- &-  & 74.8            &77.2 &87.4  &90.5 &52.3\\ 
BoT~\cite{luo2019trick}    & CVPRW19  & 94.5 &- &-  & 85.9                 &86.4 &- &-    &76.4         &- &- &-  &- \\ 
PCB~\cite{sun2018beyond}   & ECCV18   & 93.8 &- &- & 81.6                  &83.3 &- &-   &69.2           &68.2 &- &- &40.4\\ 
\hline
\multicolumn{13}{l}{\textbf{Unsupervised}} \\
ECN \cite{zhong2019invariance}  & CVPR19      & 75.1 & 87.6 & 91.6 & 43.0                 & 63.3 & 75.8 & 80.4 & 40.4    & 30.2  & 41.5 & 46.8& 10.2 \\
AE~\cite{ding2019adaptive}	& Arxiv19	& 81.6 & 91.9 & 94.6  & 58.0                     & 67.9 & 79.2 & 83.6	& 46.7 		& 32.3 & 44.4 & 50.1 & 11.7 \\
BUC~\cite{lin2019aBottom}	& AAAI19        & 66.2 & 79.6 & 84.5  & 38.3                             & 47.4 & 62.6 & 68.4 & 27.5			& - & - & - & - \\
UGA~\cite{wu2019graph}		&  ICCV19         &87.2&-&-  & 70.3                                      & 75.0 &- & - 	& 53.3	          	& 49.5 & - &- & 21.7 \\ 
\hline
\multicolumn{13}{l}{\textbf{Intra-camera supervised}} \\
MTML \cite{zhu2019intra}          & ICCVW19               & 85.3  & -    & 96.2   & 65.2                    & 71.7 &  -   & 86.9  & 50.7        & 44.1 & -   & 63.9  & 18.6 \\
PCSL \cite{qi2019progressive}     & TCSVT20                  & 87.0 & 94.8 & 96.6  & 69.4                  & 71.7 & 84.7 & 88.2 & 53.5      & 48.3   & 62.8  & 68.6& 20.7\\
ACAN~\cite{qi2019intra}           & Arxiv19                            &73.3   &87.6  &91.8 &50.6                      & 67.6 &81.2  &85.2& 45.1          &33.0   &48.0  &54.7 &12.6 \\
MATE \cite{zhu2020intra}          & Arxiv20                           & 88.7 & - & 97.1 & 71.1                           & 76.9 & - & 89.6& 56.6                & 46.0 & - & 65.3& 19.1\\
Precise-ICS  & Ours                  & \textbf{93.1} & \textbf{97.8} & \textbf{98.6}	&\textbf{83.6}       & \textbf{83.6} & \textbf{92.6} & \textbf{94.7}	&\textbf{72.0}                      & \textbf{57.7} & \textbf{71.1} & \textbf{76.3} &\textbf{31.3}   \\
\hline
\end{tabular}
}
\caption{Comparison with state-of-the-art methods.}
\label{compare_SOTA_table}
\end{table*}

\subsubsection{Effectiveness of The Inter-camera Learning Part}
After the comparisons of all intra-camera learning components, we validate the effectiveness of the inter-camera learning part. Once getting pseudo labels, we have two options for inter-camera learning. They are, $\mathcal{M}_6$ that uses a non-parametric classifier as in intra-camera learning but is camera-agnostic, and $\mathcal{M}_7$ that adopts the parametric classifier as introduced in Section~\ref{sec:model_updating}. The results in Table~\ref{ablation_table} demonstrate that the model $\mathcal{M}_7$ using the parametric classifier performs much better than the one $\mathcal{M}_6$ using the non-parametric variant. The reason we conjecture is that, the non-parametric version has much less parameters so that it is easier to get overfitting when most IDs contain abundant samples after ID association.

In addition, Table~\ref{ablation_table} also provides the results obtained by training the inter-camera learning branch in our network with entire ground-truth labels, which in essence is the fully supervised counterpart model $\mathcal{M}_8$. This model indicates the upper bound performance that can be achieved by our Re-ID network architecture. 

To investigate how the proposed graph-based ID association performs, we provide the precision and recall of the associated ID pairs using our ID association strategy. As shown in Table \ref{table_prec_rec}, the precision maintains a quite high value on both Market and Duke, showing the reliability of associated pairs for model learning. Besides, the recall is over $74\%$, indicating that most positive pairs are associated by our algorithm.

\subsubsection{Visualization of Learned Representations}
We utilize t-SNE~\cite{vanDerMaaten2008} to visualize the feature representations learned by our model components. Figure~\ref{tsne_visualization} presents the features of images in 20 IDs, respectively, learned by the camera-specific non-parametric classifiers ($\mathcal{M}_3$), the entire intra-camera learning model ($\mathcal{M}_5$), and the full model ($\mathcal{M}_7$). As Figure~\ref{tsne_visualization} shows, the camera-specific non-parametric classifiers ($\mathcal{M}_3$) gain considerable discrimination ability, but may mix up a number of difficult ID classes such as $\#9$ and $\#11$, $\#8$ and $\#13$, $\#1$ and $\#5$, as well as $\#12$ and $\#14$. With the additional quintuplet loss, the model $\mathcal{M}_5$ better separates these hard ID pairs that have very similar appearances, as sample images of $\#9$ and $\#11$ shown in the bottom row of Figure~\ref{tsne_visualization}. The full model $\mathcal{M}_7$, which incorporates both intra- and inter-camera learning, improves the intra-class compactness and inter-class separability further. The full model may still mix up a small number of IDs, but a part of the reason is because of the labeling noise, as the presented examples of $\#12$ and $\#14$.

%


\subsection{Comparison with the State-of-the-Arts}
In this section, we compare our approach (named as Precise-ICS) with all existing ICS person Re-ID methods, including MTML~\cite{zhu2019intra}, PCSL~\cite{qi2019progressive}, ACAN~\cite{qi2019intra}, and MATE~\cite{zhu2020intra}. 
 From the results we observe that the proposed approach outperforms the other ICS methods by a great margin. More specifically, the mAP is 12.5\%, 15.4\%, and 10.6\% higher and the Rank-1 accuracy is 4.4\%, 6.7\%, and 9.4\% higher than the best performances obtained by other methods on Market1501, DukeMTMC-ReID, and MSMT17 respectively. Even the model using only the intra-camera learning part (Precise-ICS: $\mathcal{M}_5$ shown in Table~\ref{ablation_table}) performs better than existing ICS methods, indicating that the proposed components in our intra-camera learning can exploit per-camera labels much more thoroughly.

In addition, we also compare our work with state-of-the-art methods under different supervision settings, including four fully supervised methods and four unsupervised methods. As expected, our approach achieves much higher performance than the unsupervised methods no matter if they transfer knowledge from extra datasets or not. Meanwhile, our approach is better than an effective supervised method PCB~\cite{sun2018beyond}, and is even comparable to recent supervised methods (DGNet~\cite{zheng2019dgnet}, BoT~\cite{luo2019trick}) on Market1501 and DukeMTMC-ReID. The results demonstrate the potential for the ICS Re-ID task to achieve high Re-ID performance while dramatically reduce labeling cost, making this supervision setting more scalable to real-world applications.   

\section{Conclusion}
In this paper, we have proposed a new approach to address the person Re-ID problem under intra-camera supervision. The proposed network consists of a shared feature extraction backbone, together with two branches for intra- and inter-camera learning respectively. According to the per-camera labeling nature of ICS, we propose jointly learned camera-specific non-parametric classifiers and a hybrid mining quintuplet loss for intra-camera learning. The designed components exploit per-camera labels thoroughly so that our intra-camera learning part only performs better than most existing ICS methods. Benefited from the discrimination ability gained in this part, the inter-camera learning module can boost the Re-ID performance further by mining ID relationship across cameras. Our full model outperforms all ICS methods by a large margin, greatly reducing the gap to the fully supervised counterparts.

{\small
\bibliographystyle{ieee_fullname}
\bibliography{reference}

\begin{thebibliography}{10}\itemsep=-1pt

\bibitem{chen2019abd}
Tianlong Chen, Shaojin Ding, Jingyi Xie, Ye Yuan, Wuyang Chen, Yang Yang, Zhou
  Ren, and Zhangyang Wang.
\newblock Abd-net: Attentive but diverse person re-identification.
\newblock In {\em ICCV}, 2019.

\bibitem{chen2018deepa}
Yanbei Chen, Xiatian Zhu, and Shaogang Gong.
\newblock Deep association learning for unsupervised video person
  re-identification.
\newblock In {\em BMVC}, 2018.

\bibitem{chen2018memory}
Yanbei Chen, Xiatian Zhu, and Shaogang Gong.
\newblock Semi-supervised deep learning with memory.
\newblock In {\em ECCV}, 2018.

\bibitem{deng2018similarity}
Weijian Deng, Liang Zheng, Qixiang Ye, Yi Yang, and Jianbin Jiao.
\newblock Image-image domain adaptation with preserved self-similarity and
  domain-dissimilarity for person reidentificatio.
\newblock In {\em CVPR}, 2018.

\bibitem{ding2019adaptive}
Yuhang Ding, Hehe Fan, Mingliang Xu, and Yi Yang.
\newblock Adaptive exploration for unsupervised person re-identification.
\newblock {\em arXiv preprint arXiv:1907.04194}, 2019.

\bibitem{unsup_clustering}
Hehe Fan, Liang Zheng, Chenggang Yan, and Yi Yang.
\newblock Unsupervised person re-identification: Clustering and fine-tuning.
\newblock {\em ACM Transactions on Multimedia Computing, Communications, and
  Applications}, 14(4):83, 2018.

\bibitem{he2016deep}
Kaiming He, Xiangyu Zhang, Shaoqing Ren, and Jian Sun.
\newblock Deep residual learning for image recognition.
\newblock In {\em CVPR}, 2016.

\bibitem{hermans2017defense}
Alexander Hermans, Lucas Beyer, and Bastian Leibe.
\newblock In defense of the triplet loss for person re-identification.
\newblock {\em arXiv preprint arXiv:1703.07737}, 2017.

\bibitem{Kingma2014}
Diederik~P. Kingma and Jimmy~Lei Ba.
\newblock Adam : A method for stochastic optimization.
\newblock {\em arXiv preprint arXiv:2014}, 2014.

\bibitem{kodirov2016person}
Elyor Kodirov, Tao Xiang, Zhenyong Fu, and Shaogang Gong.
\newblock Person re-identification by unsupervised $l_1$ graph learning.
\newblock In {\em European conference on computer vision}, pages 178--195.
  Springer, 2016.

\bibitem{krizhevsky2012imagenet}
Alex Krizhevsky, Ilya Sutskever, and Geoffrey~E Hinton.
\newblock Imagenet classification with deep convolutional neural networks.
\newblock In {\em Advances in neural information processing systems}, pages
  1097--1105, 2012.

\bibitem{lin2019aBottom}
Yutian Lin, Xuanyi Dong, Liang Zheng, Yan Yan, and Yi Yang.
\newblock A bottom-up clustering approach to unsupervised person
  re-identification.
\newblock In {\em AAAI}, 2019.

\bibitem{luo2019trick}
Hao Luo, Youzhi Gu, Xingyu Liao, Shenqi Lai, and Wei Jiang.
\newblock Bag of tricks and a strong baseline for deep person
  re-identification.
\newblock In {\em CVPRW}, 2019.

\bibitem{patwary2010experiments}
Md~Mostofa~Ali Patwary, Jean Blair, and Fredrik Manne.
\newblock Experiments on union-find algorithms for the disjoint-set data
  structure.
\newblock In {\em International Symposium on Experimental Algorithms}, pages
  411--423. Springer, 2010.

\bibitem{qi2019intra}
Lei Qi, Lei Wang, Jing Huo, Yinghuan Shi, and Yang Gao.
\newblock Adversarial camera alignment network for unsupervised cross-camera
  person re-identification.
\newblock {\em arXiv preprint arXiv:1908.00862}, 2019.

\bibitem{qi2019progressive}
Lei Qi, Lei Wang, Jing Huo, Yinghuan Shi, and Yang Gao.
\newblock Progressive cross-camera soft-label learning for semi-supervised
  person re-identification.
\newblock {\em arXiv preprint arXiv:1908.05669}, 2019.

\bibitem{ristani2016performance}
Ergys Ristani, Francesco Solera, Roger Zou, Rita Cucchiara, and Carlo Tomasi.
\newblock Performance measures and a data set for multi-target, multi-camera
  tracking.
\newblock In {\em ECCV}, 2016.

\bibitem{Sohn2016Npair}
Kihyuk Sohn.
\newblock Improved deep metric learning with multi-class n-pair loss objective.
\newblock In {\em NIPS}, 2016.

\bibitem{sun2018beyond}
Yifan Sun, Liang Zheng, Yi Yang, Qi Tian, and Shengjin Wang.
\newblock Beyond part models: Person retrieval with refined part pooling (and a
  strong convolutional baseline).
\newblock In {\em ECCV}, pages 480--496, 2018.

\bibitem{vanDerMaaten2008}
Laurens van~der Maaten and Geoffrey Hinton.
\newblock Visualizing data using {t-SNE}.
\newblock {\em Journal of Machine Learning Research}, 9:2579--2605, 2008.

\bibitem{wei2018person}
Longhui Wei, Shiliang Zhang, Wen Gao, and Qi Tian.
\newblock Person transfer gan to bridge domain gap for person
  re-identification.
\newblock In {\em CVPR}, 2018.

\bibitem{wu2019graph}
Jinlin Wu, Yang Yang, Hao Liu, Shengcai Liao, Zhen Lei, and Stan~Z. Li.
\newblock Unsupervised graph association for person re-identification.
\newblock In {\em ICCV}, 2019.

\bibitem{wu2018memory}
Zhirong Wu, Yuanjun Xiong, Stella~X. Yu, and Dahua Lin.
\newblock Unsupervised feature learning via non-parametric instance
  discrimination.
\newblock In {\em CVPR}, 2018.

\bibitem{xiao2017memory}
Tong Xiao, Shuang Li, Bochao Wang, Liang Lin, and Xiaogang Wang.
\newblock Joint detection and identification feature learning for person
  search.
\newblock In {\em CVPR}, 2017.

\bibitem{yangqi2019semi}
Qize Yang, Ancong Wu, and Wei-Shi Zheng.
\newblock Deep semi-supervised person re-identification with external memory.
\newblock In {\em ICME}, 2019.

\bibitem{yang2019patch}
Qize Yang, Hong-Xing Yu, Ancong Wu, and Wei-Shi Zheng.
\newblock Patch-based discriminative feature learning for unsupervised person
  re-identification.
\newblock In {\em CVPR}, pages 3633--3642, 2019.

\bibitem{Zhai2019loss}
Yao Zhai, Xun Guo, Yan Lu, and Houqiang Li.
\newblock In defense of the classification loss for person re-identification.
\newblock In {\em CVPRW}, 2019.

\bibitem{zhang2019dsa}
Zhizheng Zhang, Cuiling Lan, Wenjun Zeng, and Zhibo Chen.
\newblock Densely semantically aligned person re-identification.
\newblock In {\em CVPR}, 2019.

\bibitem{7410490}
L. Zheng, L. Shen, L. Tian, S. Wang, J. Wang, and Q. Tian.
\newblock Scalable person re-identification: A benchmark.
\newblock In {\em ICCV}, 2015.

\bibitem{zheng2019dgnet}
Zhedong Zheng, Xiaodong Yang, Zhiding Yu, Liang Zheng, Yi Yang, and Jan Kautz.
\newblock Joint discriminative and generative learning for person
  re-identification.
\newblock In {\em CVPR}, 2019.

\bibitem{zheng2017unlabeled}
Zhedong Zheng, Liang Zheng, and Yi Yang.
\newblock Unlabeled samples generated by gan improve the person
  re-identification baseline in vitro.
\newblock In {\em ICCV}, 2017.

\bibitem{zhong2019invariance}
Zhun Zhong, Liang Zheng, Zhiming Luo, Shaozi Li, and Yi Yang.
\newblock Invariance matters: Exemplar memory for domain adaptive person
  re-identification.
\newblock In {\em CVPR}, 2019.

\bibitem{zhong2019memory}
Zhun Zhong, Liang Zheng, Zhiming Luo, Shaozi Li, and Yi Yang.
\newblock Learning to adapt invariance in memory for person re-identification.
\newblock {\em arXiv preprint arXiv:1908.00485}, 2019.

\bibitem{zhou2019osnet}
Kaiyang Zhou, Yongxin Yang, Andrea Cavallaro, and Tao Xiang.
\newblock Omni-scale feature learning for person re-identification.
\newblock In {\em ICCV}, 2019.

\bibitem{zhu2020intra}
Xiangping Zhu, Xiatian Zhu, Minxian Li, Pietro Morerio, Vittorio Murino, and
  Shaogang Gong.
\newblock Intra-camera supervised person re-identification.
\newblock {\em arXiv preprint arXiv:2002.05046}, 2020.

\bibitem{zhu2019intra}
Xiangping Zhu, Xiatian Zhu, Minxian Li, Vittorio Murino, and Shaogang Gong.
\newblock Intra-camera supervised person re-identification: A new benchmark.
\newblock In {\em ICCVW}, 2019.

\end{thebibliography}
}

\end{document}